\documentclass[letterpaper]{IEEEtran}

\usepackage[colorlinks,urlcolor=blue,linkcolor=blue,citecolor=blue]{hyperref}

\usepackage{color,array}

\usepackage{graphicx}

\usepackage{color,array}
\usepackage{xcolor,colortbl}

\definecolor{Gray}{gray}{0.85}
\newcolumntype{a}{>{\columncolor{Gray}}c}

\usepackage{graphicx} 
\usepackage{float}
\usepackage{subcaption}

\usepackage{algorithm}
\usepackage{algpseudocode}

\usepackage{amsmath}
\DeclareMathAlphabet\mathbfcal{OMS}{cmsy}{b}{n}
\usepackage{bm}

\usepackage{multirow}

\begin{document}

\title{A Human-in-the-Middle Attack against Object Detection Systems} 

\author{
    Han Wu, Sareh Rowlands and Johan Wahlstr\"om \thanks{ Han Wu, Sareh Rowlands and Johan Wahlstr\"om are currently with the University of Exeter, Stocker Rd, Exeter EX4 4PY UK (e-mail: hw630@exeter.ac.uk; s.rowlands@exeter.ac.uk; j.wahlstrom@exeter.ac.uk).}
}

\markboth{A Human-in-the-Middle Attack against Object Detection}
{H. Wu \MakeLowercase{\textit{et al.}}: A Human-in-the-Middle Attack against Object Detection}

\maketitle

\begin{abstract}
Object detection systems using deep learning models have become increasingly popular in robotics thanks to the rising power of CPUs and GPUs in embedded systems. However, these models are susceptible to adversarial attacks. While some attacks are limited by strict assumptions on access to the detection system, we propose a novel hardware attack inspired by Man-in-the-Middle attacks in cryptography. This attack generates a Universal Adversarial Perturbations (UAP)  and injects the perturbation between the USB camera and the detection system via a hardware attack. Besides, prior research is misled by an evaluation metric that measures the model accuracy rather than the attack performance. In combination with our proposed evaluation metrics, we significantly increased  the strength of adversarial perturbations. These findings raise serious concerns for applications of deep learning models in safety-critical systems, such as autonomous driving.
\end{abstract}

\begin{IEEEImpStatement}
Advancements in deep neural networks have ushered in a new era of robotics, characterized by intelligent robots with a comprehensive understanding of the environment, thanks to deep learning models. However, it is no more a secret that deep learning models are vulnerable to adversarial attacks. Besides existing digital and physical attacks, we introduce a novel 'Human-in-the-Middle' hardware attack that injects digital perturbation into the physical sensor. Our research opens up new possibilities for adversarial attacks, and we hope to embrace deep learning models securely for robotic applications.


\end{IEEEImpStatement}


\section{Introduction}

The development of deep neural networks has enabled the creation of intelligent robots that possess a more comprehensive perception of the environment than traditional robots. However, this shift towards intelligent robots has also brought with it an increasing risk of adversarial attacks, especially in safety-critical applications.  It has been a decade since the existence of adversarial examples was first identified by Biggio et al. \cite{biggio2013evasion}, and Szegedy et al. \cite{szegedy2013intriguing}, in which they fooled an image classification model by adding a small perturbation to the input image.  Although the perturbation was imperceptible to humans, it caused the deep learning model to produce erroneous classification results. The attack was later extended from classification models to detection models \cite{han2023detection, LuSFF17}. 


Adversarial attacks against deep learning models can be divided into two categories: digital attacks and physical attacks. Digital attacks directly apply perturbations to the digital input image by modifying pixel values \cite{han2023driving}, while physical attacks involve printing the perturbation on physical objects such as posters \cite{lee2019physical} or T-shirts \cite{xu2020adversarial}.

However, both digital and physical attacks have their limitations. Digital perturbation requires access to the detection system, making it difficult to apply in real-world scenarios such as hacking into a self-driving car. Physical attacks, on the other hand, are sensitive to position and angle variations. For instance, experiments in \cite{LuSFF17} showed that an autonomous vehicle only misclassified traffic signs placed within 0.5 meters of the camera and viewed from specific angles. Moreover, these attacks lack flexibility, as once the adversarial object is printed, it can only be changed through reprinting. The trial-and-error process of finding a successful attack object can take a long time and require significant amounts of printing.

\begin{figure}[tpb]
    \centering
    \begin{subfigure}[b]{\linewidth}
        \includegraphics[width=\linewidth]{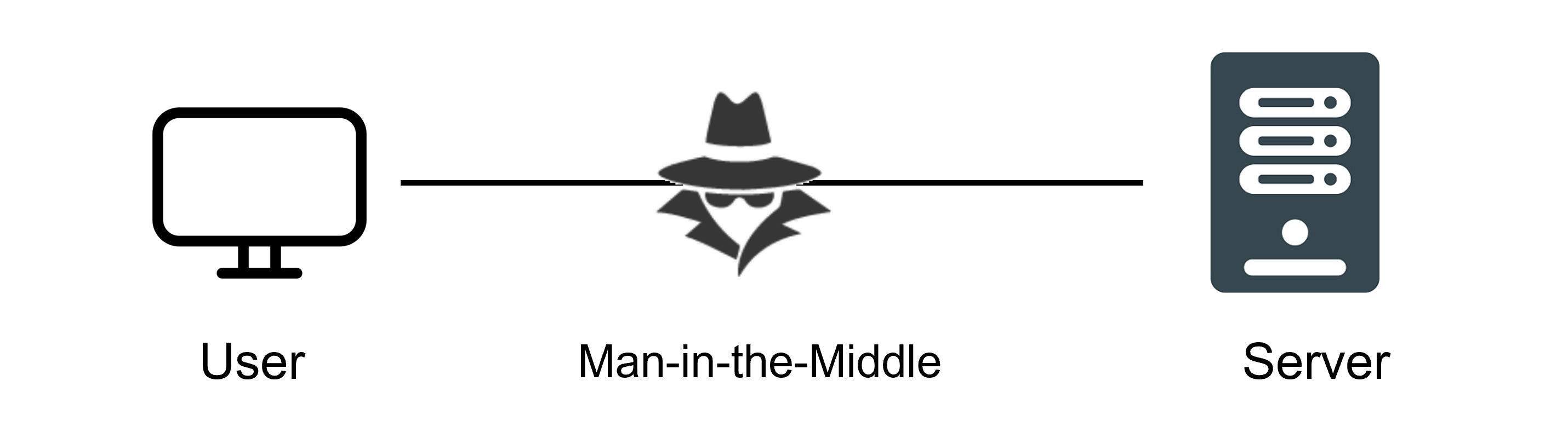}
        \caption{Man-in-the-Middle  Attack in Network Security}
        \label{fig:mitm} 
    \end{subfigure}

    \begin{subfigure}[b]{\linewidth}
        \includegraphics[width=\linewidth]{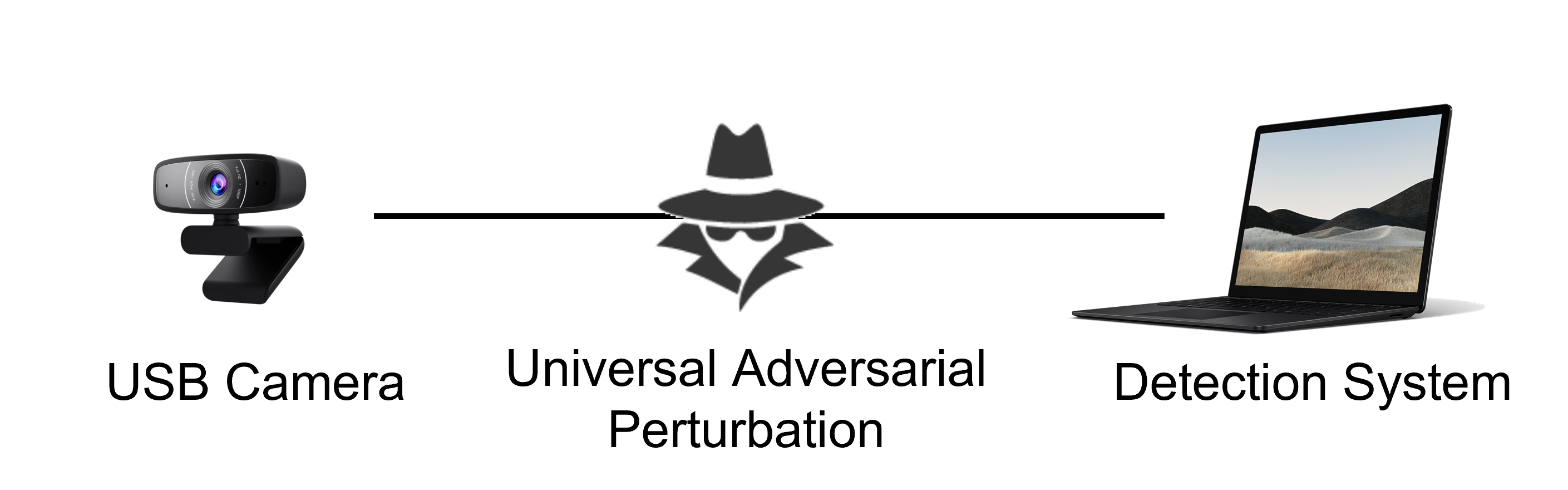}
        \caption{Human-in-the-Middle  Attack in Deep Learning}
        \label{fig:minm}
    \end{subfigure}

    \begin{subfigure}[b]{\linewidth}
        \includegraphics[width=\linewidth]{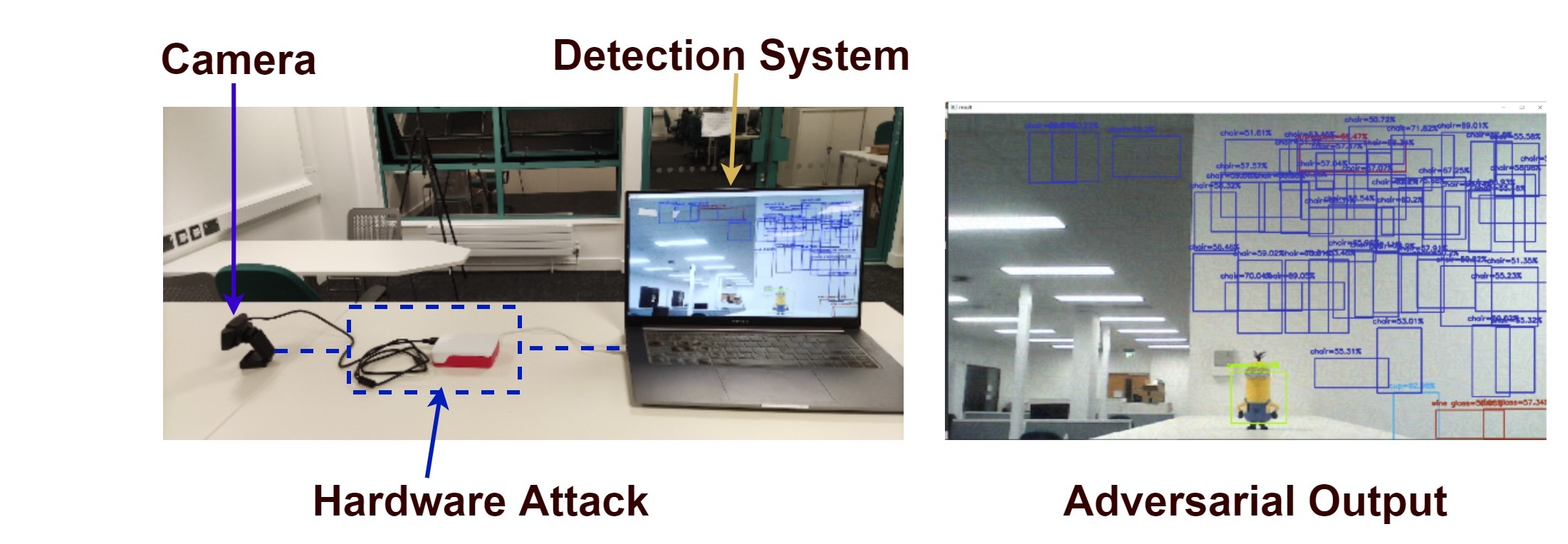}
        \caption{Demo of the Hardware Attack}
        \label{fig:demo}
    \end{subfigure}

  \caption{The Human-in-the-Middle (HitM)  hardware attack.}
  \label{fig:overview}
\end{figure}

\subsection{Contributions}

This paper presents a novel hardware attack that combines the flexibility of physical attacks with the efficiency of digital attacks, inspired by Man-in-the-Middle Attacks in network security (refer to Fig. \ref{fig:mitm}). In this attack, the adversary intercepts and manipulates the image data transmitted between a USB camera and a detection system (refer to Figs. \ref{fig:minm} and \ref{fig:demo}). 

The key contributions of this research are summarized as follows:

\begin{enumerate}
    \item We present a novel hardware attack, called Human-in-the-Middle  attack, that offers both efficiency and ease of deployment for adversarial attacks\footnote{The source code of the hardware attack is available on GitHub: \url{https://github.com/wuhanstudio/adversarial-camera}}. By utilizing learning rate decay during the generation of the perturbation, our attack is capable of generating more bounding boxes than competing attack methods.
    \item We introduce three new evaluation metrics that offer a more nuanced approach to evaluating adversarial attacks. Unlike existing metrics that make a binary decision for each bounding box, our metrics consider the confidence value and probability vector in a linear fashion.
    \item We devise and open source the white-box adversarial toolbox\footnote{The source code of the toolbox is available on GitHub: \url{https://github.com/wuhanstudio/whitebox-adversarial-toolbox}} that simplifies the process of generating adversarial perturbations. The toolbox focuses on real-time white-box attacks against object detection models.
\end{enumerate}


\section{Preliminaries}


\subsection{Object Detection Models}

The task of object detection aims to locate the position and classify the category of each object in an image. Therefore, the task consists of two distinct problems: localization and classification. Existing object detection models can be categorized into two types, one-stage and two-stage methods, based on whether these two problems are solved together or separately \cite{Zhao2019}. Two of the most widely deployed one-stage models are YOLO \cite{redmon2016you, redmon2018yolov3, bochkovskiy2020yolov4} and SSD \cite{liu2016ssd}, which can achieve real-time performance on CPUs without GPUs. Faster RCNN \cite{ren2015faster} and Mask RCNN \cite{he2017mask} are two well-known two-stage models. 



In robotic applications, one-stage models are generally preferred due to their speed and acceptable accuracy in most situations. In this study, we investigate how these attacks affect real-time robotic applications and focus on energy-efficient one-stage models.



\subsection{Adversarial Attacks}

 The Fast Gradient Sign Method (FGSM) \cite{GoodfellowSS14} was the first adversarial attack against classification models that uses gradients of deep neural networks to generate image-specific perturbations, which is more efficient than optimization-based methods proposed by Biggio et al. \cite{biggio2013evasion}, and Szegedy et al. \cite{szegedy2013intriguing}. 

However, for real-world robotic applications, it is more practical to use Universal Adversarial Perturbations (UAPs) \cite{moosavi2017universal, li2021universal, wu2023adversariald}, which are image-agnostic. UAPs demonstrated the ability to fool classification models on most images in a dataset using a single perturbation. Adversarial attacks have since been extended from image classification to detection models \cite{gurbaxani2018traits, han2023detection}.

In addition to image-specific and image-agnostic methods, it is also possible to classify adversarial attacks into data-driven and data-independent approaches.  Data-driven approaches require access to the input image, while data-independent methods do not need access to the input data. Both approaches may need access to the parameters and the architecture of the target model, depending on whether they are white-box attacks or black-box attacks.  Generally, data-driven approaches achieve a higher fooling rate as they have more information at their disposal. Data-driven methods include gradient-based methods \cite{chow2020adversarial,li2021universal, xie2017adversarial, wu2023adversarial}, methods using Generative Adversarial Networks (GANs) \cite{hashemi2020transferable,Wei2019}, and optimization-based methods \cite{carlini2017towards,liao2021transferable}. 







To deploy the UAPs generated using the aforementioned methods, Wang et al. categorized existing physical attacks into invasive attacks and
non-invasive attacks \cite{wang2022survey}. Invasive attacks deploy the perturbation by attaching a patch \cite{lee2019physical, xu2020adversarial, hu2021naturalistic, thys2019fooling} or changing the texture \cite{wang2021dual, wang2022fca} of the target object. For example, by adding extra constraints, such as the Sub-sampled Non-Printability Score (SNPS), to the loss function of data-driven methods, we can generate physical perturbations that preserve the adversarial effect if printed out on a poster \cite{lee2019physical}. On the other hand, non-invasive attacks perform physical attacks by changing the environment rather than the target object. For example, Gnanasambandam et al. fool classification models by changing the illumination of the environment using a low-cost projector \cite{gnanasambandam2021optical}, and Zhong et al. exploit the shadow to fool deep learning models \cite{zhong2022shadows}.



\begin{figure}[bt]
    \centering
    \includegraphics[width=0.9\linewidth]{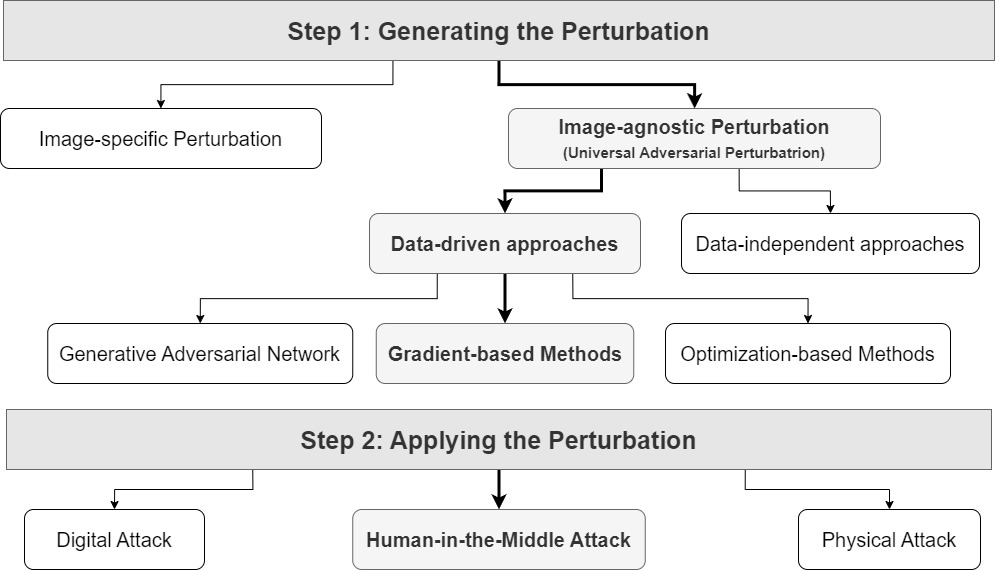}
    \caption{Our approach to generate and apply adversarial perturbations in bold.}
    \label{fig:attack}
\end{figure}

We propose a novel Human-in-the-Middle (HitM)  hardware attack that neither modifies the target object (invasive attacks) nor changes the illumination or shadow of the environment (non-invasive attacks). Instead, our attack injects the perturbation into the physical communication channel, combining the advantages of digital and physical attacks. Wang et al. propose a Main-in-the-Middle attack against Machine-Learning-as-a-service applications that exploit vulnerabilities in networking to stealthy manipulate the submitted data, which is a traditional network attack \cite{wang2020man}. In a parallel research endeavor, Liu et al. focus on local-search-based black-box attacks against image classification models and inject adversarial strips to a MIPI camera using an FPGA \cite{liu2022practical}. On the other hand, we focus on gradient-based white-box attacks against object detection models and inject universal adversarial perturbations to a USB camera via an ARM Linux board, as depicted in Fig. \ref{fig:attack}. Due to hardware limitations, Liu's method can only inject monochrome adversarial strips, while we can inject polychrome adversarial perturbations. 




\clearpage

\section{The Human-in-the-Middle Attack}

This section introduces the PCB attack, a novel gradient-based method designed to generate image-agnostic UAPs. The name "PCB" comes from the fact that the output of the object detection model is separated into three components: probability vector (P), confidence value (C), and bounding boxes (B). The perturbation is then applied using a hardware attack. The acronym PCB is fitting for a hardware attack, as it is also used to refer to Printed Circuit Boards (PCB).

\subsection{Problem Formulation} 


In Section II, we discussed that object detection models can be categorized into one-stage models (e.g., YOLO, SSD) and two-stage models (e.g., Faster-RCNN, Mask-RCNN). Despite the differences in their structures, all these models share common inputs and outputs. To describe these inputs and outputs, we introduce the following mathematical notation:

\begin{itemize}
    \item $x$: The original clean input image.
    \item $\delta$: The adversarial perturbation.
    \item $x^{'}$: The adversarial input image $x^{'} = x + \delta$.
    \item $K$: The total number of candidate classes.
    \item $N$: The total number of candidate bounding boxes.
    \item $\mathcal{O}(x)$: The output of $N$ candidate bounding boxes from the model given the input image $x$. 
    \item $o_i(x)$: The $i_{th}$ output in $\mathcal{O}(x) = \{o_1, o_2, o_3, ..., o_N\}$, where $o_i=(b_i, c_i, p_i)$. $1 \leq i \leq N$.
    \item $b_i$: The location and dimension of the $i_{th}$ candidate box. $b_i=(b^x_i, b^y_i, b^w_i, b^h_i)$ represents a bounding box at position $(b^x_i, b^y_i)$ with width $b^w_i$ and height $b^h_i$,
    \item $c_i$: The confidence value (objectness) of the $i_{th}$ candidate box that represents how probable it is that the the candidate box represents an object.
    \item $p_i$: The softmax probability vector of the $i_{th}$ candidate box. $p_i=(p^1_i, p^2_i, ..., p^K_i)$ for $K$ classes and $\sum{p_i}=1$.
\end{itemize}

Given an input image $x$, the object detection model outputs $N$ candidate bounding boxes $\mathcal{O}(x) = \{o_1, o_2, o_3, ..., o_N\}$. Each candidate box $o_i=(b_i, c_i, p_i)$ contains $b_i=(b^x_i, b^y_i, b^w_i, b^h_i)$ that represents the location and dimension of the box, the confidence value $c_i \in [0, 1]$ that represents how probable it is that the the candidate box represents an object, and the softmax probability vector, $p_i=(p^1_i, p^2_i, ..., K_i)$ for $K$ classes. The raw outputs from the detection model $\mathcal{O}(x)$ may contain several thousand candidate bounding boxes. We then use the Non-maximum Suppression (NMS) method \cite{bodla2017soft} to filter out bounding boxes with low confidence values, and high Intersection over Union (IoU) to generate final detection results.


An adversarial example $x^{'} = x + \delta$ aims to fool the detection model by making it output candidate boxes $\mathcal{O}(x^{'})$ that are different from the candidate boxes $\mathcal{O}(x)$ outputted by the model for the original input image $x$. For example, the adversarial output $\mathcal{O}(x^{'})$ may detect more false positive objects after the non-maximum suppression (NMS) process. To achieve this, we need to generate a perturbation $\delta$ that can be added to the original image $x$ to produce the adversarial image $x^{'}$. In the following subsections, we will describe how to generate the perturbation $\delta$ using the proposed PCB attack.

\subsection{Generating the perturbation (The PCB Attack)}


Gradient-based methods use a similar approach to generate both image-specific and image-agnostic perturbations. For image-specific perturbations, given an input image, the method iterates over a single image to produce the perturbation. 
For image-agnostic perturbations, given the entire dataset, the method iterates over multiple images to generate the Universal Adversarial Perturbation (UAP). We will first describe how we generate the image-specific perturbations and then extend the attack to generate the image-agnostic perturbations.

\subsubsection{\textbf{Image-specific PCB Attack}}

The intuition behind gradient-based methods is straightforward. During the training process, we minimize the training loss 
\begin{equation}
\min_{\mathcal{W}} \ \mathcal{L}_{train} = f(\mathcal{W}; x, \mathcal{O})
\end{equation}
by updating the model weights. Note that the training loss is a function of the input $x$, the model weights $\mathcal{W}$, and the ground truth $\mathcal{O}$. 

However, our objective is to fool the detection model to make inaccurate predictions. Therefore, during the attack, we maximize the adversarial loss 
\begin{equation}
\label{eq_adversarial_opt}
\max_{x} \ \mathcal{L}_{adv} = f(x; \mathcal{O}^{\ast}, \mathcal{W})
\end{equation}
by updating the input $x$ and using the desired adversarial outputs $\mathcal{O}^{\ast}$. Different gradient-based methods use different adversarial loss functions $\mathcal{L}_{adv}$ and construct desired adversarial outputs $\mathcal{O}^{\ast}$ differently. In our attack, we separate the Probability vector and Confidence value (PC) with Bounding boxes (B) and investigate the two adversarial loss functions
\begin{equation}
\mathcal{L}_{PC}(x) = \sum{\sigma(c_i) * \sigma(p_i)}
\end{equation}
and
\begin{equation}
\mathcal{L}_{PCB}(x) = \frac{\sum{(\sigma(c_i) * \sigma(p_i)}}{\sum{[\sigma(w_i) * \sigma(h_i)]^2}}.
\end{equation}
where $\sigma(\cdot)$ is the sigmoid function. 

By maximizing the adversarial loss ( $\mathcal{L}_{PCB}(x)$ and $\mathcal{L}_{PC}(x)$), we generate large amounts of incorrect bounding boxes (fabrication attack). By minimizing the loss, we remove bounding boxes (vanishing attack).  Using $\mathcal{L}_{PCB}(x)$ gives smaller bounding boxes, while $\mathcal{L}_{PC}(x)$ gives larger ones .

The optimization of \eqref{eq_adversarial_opt} is performed by first zero-initializing the perturbation $\delta$, and then using Projected Gradient Descent (PGD) \cite{madry2017towards} with learning rate decay  in every iteration $t$ , so that 
\begin{equation}
\delta_{t+1} = proj_p(\delta_{t} + \alpha sign(\frac{\partial L_{adv}(x'_{t};O^*)}{\partial x'_{t}} )).
\end{equation}
The image-specific PCB attack is summarized in Algorithm \ref{alg:image-specific}, where 
$\texttt{proj}_{\infty}(\delta,\epsilon)$ is the projection function $\min(\delta,\epsilon)$  using $l_{\infty}$ norm , and $\texttt{clip}(-1, 1)$ is the unit clip function.

\begin{algorithm}
    \caption{Image-specific PCB Attack}\label{alg:image-specific}
    \begin{algorithmic}[1]
        \State Input: The target model, the input image $x$.
        \State Parameters: The learning rate $\alpha$, learning rate decay $k$, number of iterations $n$, and strength of the attack $\epsilon$.
        \State Output: Image-specific perturbation $\delta$
        \State Initialize $\delta \leftarrow 0$
        \For{$i = 1:n$}
            \State $x^{'} = x + \delta$  
            \State $\nabla = \frac{\partial L_{adv}^*(x';O^*)}{\partial x^{'}}$
            \State $\delta \leftarrow \delta + \alpha * \texttt{sign}(\nabla)$
            \State $\delta \leftarrow \texttt{clip}(-1, 1)$
            \State $\delta \leftarrow  \texttt{proj}_{\infty}(\delta,\epsilon)$ 
            \State // Learning Rate Decay*
            \State $\alpha = \alpha * k$ 
        \EndFor
    \end{algorithmic}
\end{algorithm}

\begin{algorithm}
    \caption{Image-agnostic PCB Attack (UAP)}\label{alg:image-agnostic}
    \begin{algorithmic}[1]
        \State Input: The target model, the sample images $X_S$.
        \State Parameters: The learning rate $\alpha$, learning rate decay $k$, number of iterations $n$, and strength of the attack $\epsilon$.
        \State Output: Image-specific perturbation $\delta$
        \State Initialize $\delta \leftarrow 0$
        \For{$i = 1:n$}
            \For{ each image $x \in X_s$}
                \State $x^{'} = x + \delta$  
                \State $\nabla = \frac{\partial L_{adv}(x^{'})}{\partial x^{'}}$
                \State $\delta \leftarrow \delta + \alpha * \texttt{sign}(\nabla)$ 
                \State $\delta \leftarrow \texttt{clip}(-1, 1)$
                \State $\delta \leftarrow  \texttt{proj}_{\infty}(\delta,\ \epsilon)$ 
            \EndFor
        \State // Learning Rate Decay*
        \State $\alpha = \alpha * k$
        \EndFor
    \end{algorithmic}
\end{algorithm}

\begin{figure*}[bthp]
    \centering
    \begin{subfigure}[b]{0.46\textwidth}
        \includegraphics[width=0.9\linewidth]{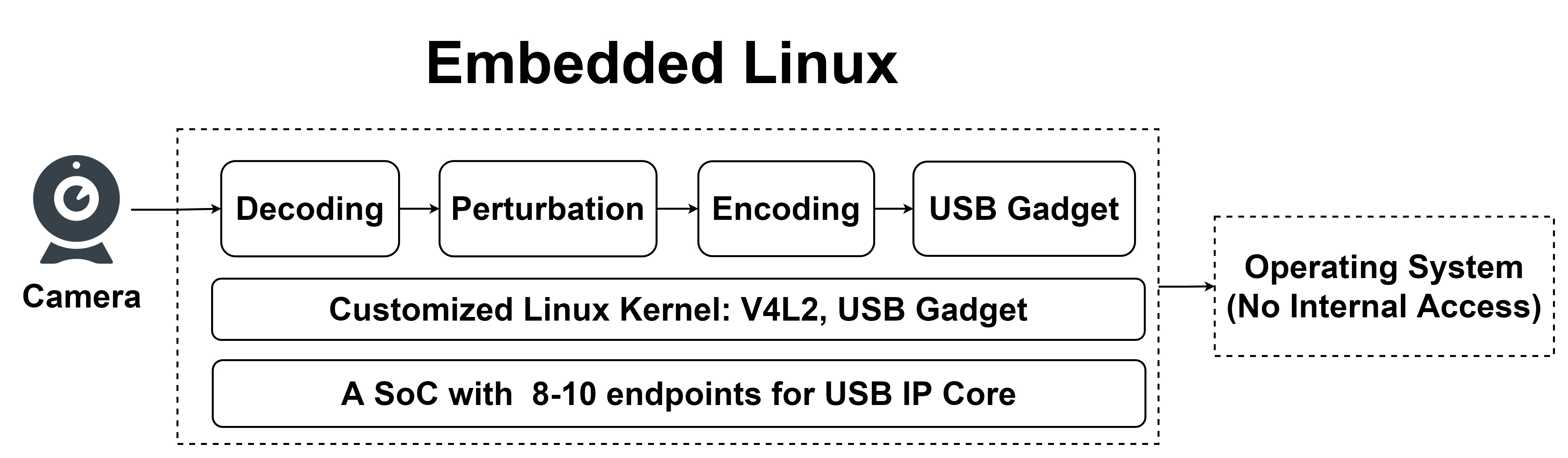}
        \caption{The Design of the Embedded System.}
        \label{fig:design}
    \end{subfigure}
    \begin{subfigure}[b]{0.46\textwidth}
        \includegraphics[width=\linewidth]{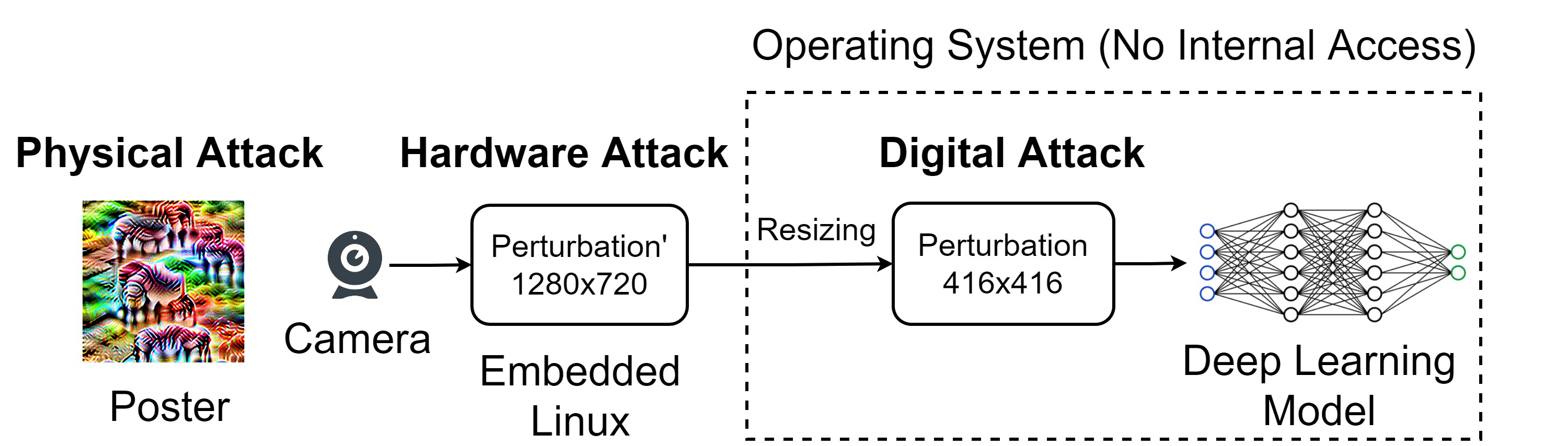}
        \caption{The Physical Attack, Hardware Attack and Digital Attack.}
        \label{fig:comparison} 
    \end{subfigure}
  \caption{The architecture of the Human-in-the-Middle  hardware attack and its differences from physical and digital attacks.}
  \label{fig:hardware}
\end{figure*}

\subsubsection{\textbf{Image-agnostic PCB Attack}}


We can extend the method to an image-agnostic attack by iterating over a collection of images $X_{s} = {x_{1}, x_{2}, ..., x_{n}}$, where $n$ is the number of available images to the attacker. $X_{s}$ can be thought of as the training set or a video clip from the target scene. Initially, we generate a random or zero-initialized perturbation $\delta$ that is of the same dimension as the input of the detection model. 

In each iteration, we update $\delta$ using the gradient of input with respect to $\mathcal{L}_{adv}$. The learning rate $\alpha$ is relatively small compared to the image-specific PCB attack to ensure that the perturbation is universal across images. We summarize the image-agnostic PCB attack in algorithm \ref{alg:image-agnostic}.


\subsection{Applying the perturbation (the hardware attack)}

In Section I, we mentioned that conducting digital attacks can be challenging due to the lack of access to the internal system. Input images are often resized and processed by intermediate components before being fed into the detection system. Therefore, an attacker needs to penetrate the operating system and inject malicious code into the embedded system. 


To address this problem, the Human-in-the-Middle  hardware attack was developed. By eavesdropping and manipulating the image data before it reaches the detection system, the perturbation can be applied without access to the operating system.  Unlike physical attacks, the hardware attack is robust to position and angle variations because the perturbation is injected by directly modifying pixel values. Thus, the perturbation won't be sheared when viewing from different angles.

To implement the hardware attack, specialized hardware such as Raspberry Pi Zero/4 or I.MX6UL is required, which can read raw images from the USB camera and then inject the perturbation (see Fig. \ref{fig:hardware}). To conceal the attack from the operating system, a virtual camera needs to be simulated to the detection system. This requires a Linux kernel that supports the V4L2 driver, the USB gadget framework, and configfs.

\begin{figure}[bp]
    \centering
    \begin{subfigure}[b]{0.46\textwidth}
        \centering
        \includegraphics[width=\linewidth]{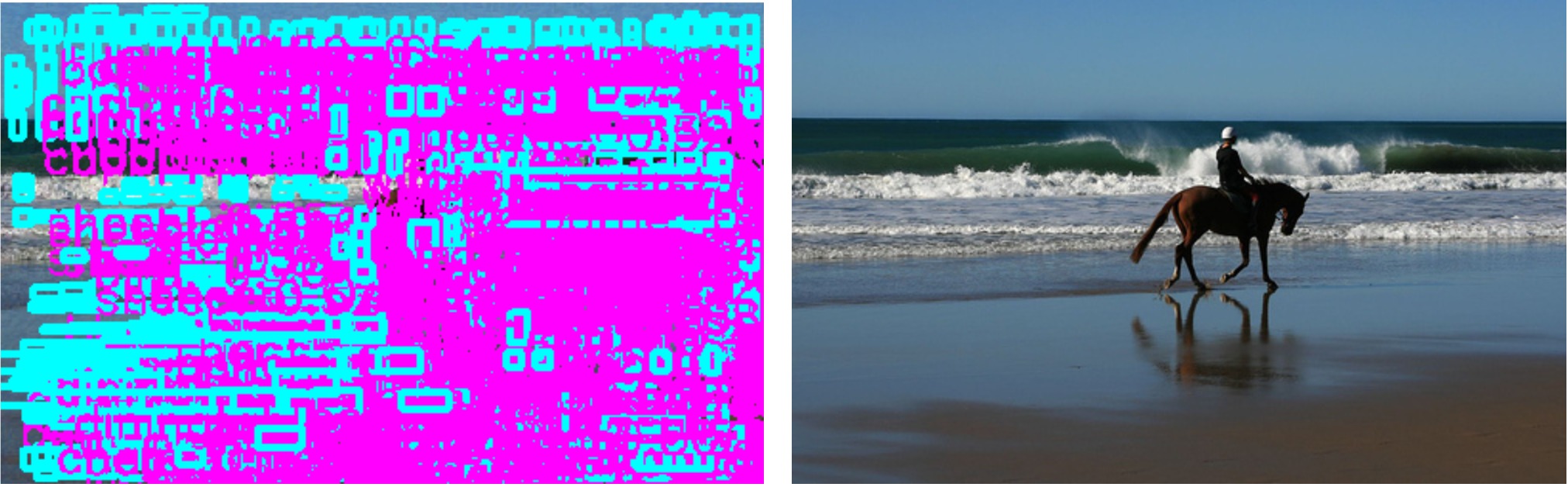}
        \caption{The mAP cannot distinguish between fabrication and vanishing attacks (both \text{mAP}=0). 
        }
        \label{fig:fabrication} 
    \end{subfigure}
    \begin{subfigure}[b]{0.46\textwidth}
        \centering
        \includegraphics[width=\linewidth]{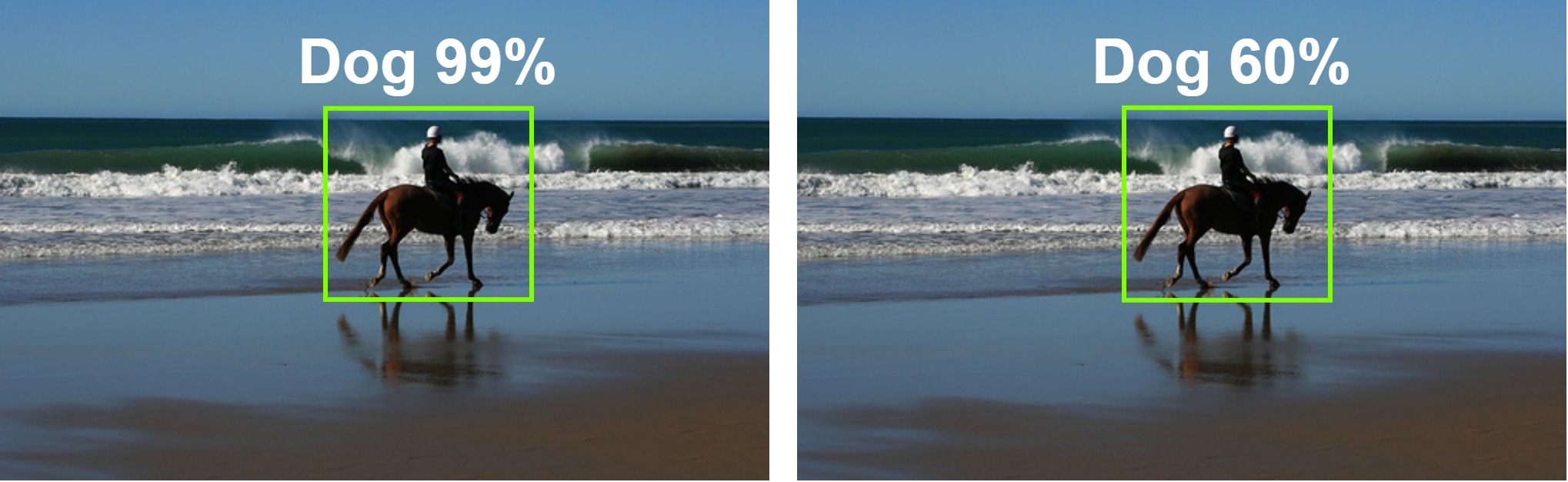}
        \caption{The mAP does not consider confidence values (both \text{mAP}=0). 
        }
        \label{fig:mislabel}
    \end{subfigure}
  \caption{The limitations of Mean Average Precision (mAP).}
  \label{fig:map}
\end{figure}

\begin{figure*}[btph]
    \centering
    \includegraphics[width=0.80\linewidth]{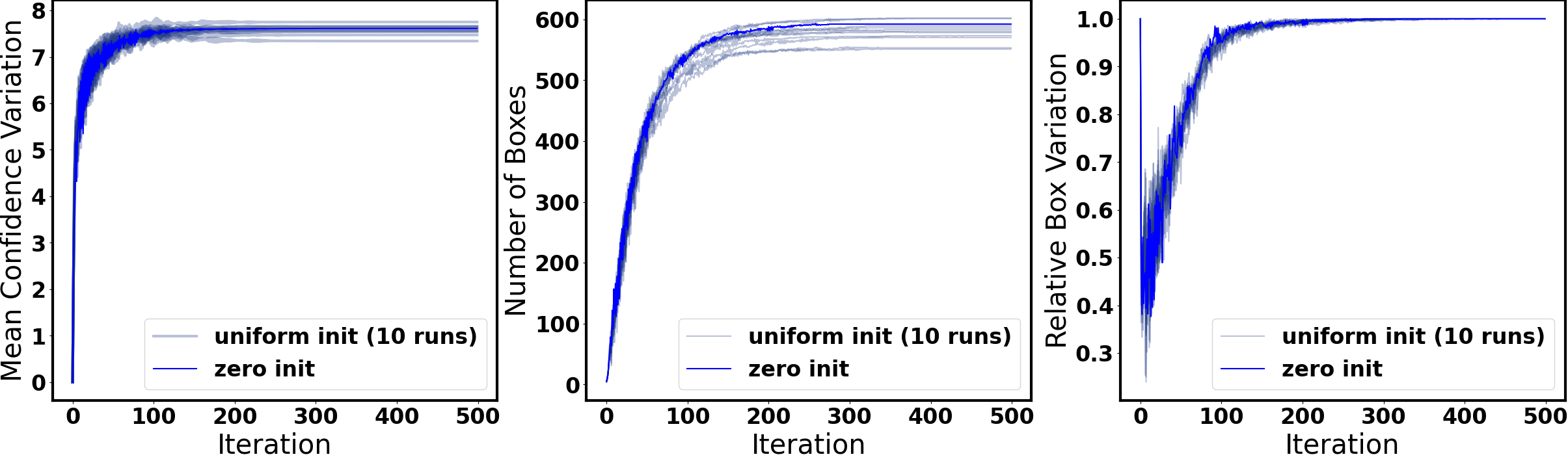}
    \caption{The image-specific  PCB fabrication attack using different initialization methods.}
    \label{fig:init}
\end{figure*}

\section{Experimental Evaluation}

This section aims to provide insight into why the Mean Average Precision (mAP) not suitable for evaluating adversarial attacks. For adversarial attacks, the choice of the adversarial loss function determines the type of attack to be conducted (e.g., fabrication or vanishing), whereas the strength of the attack is determined by the iterative optimization process. In this study, we employ our novel evaluation metrics to investigate the  iterative optimization process  and achieve more efficient attacks against a one-stage detection model, namely YOLO, on the VOC2012 \cite{pascal-voc-2012} and CARLA \cite{deschaud2021kitticarla} dataset.


\subsection{Evaluation Metrics}

The mAP \cite{cartucho2018} is typically used to both to measure the accuracy of object detection models and to evaluate the strength of adversarial attacks. However, it can be noticed that the mAP cannot distinguish between different attacks. 

For example, both the fabrication and vanishing attacks result in an mAP $\approx$ 0, even though they serve different attacking purposes (see Fig. \ref{fig:fabrication}). Similarly, while an attacker will prefer a stronger attack (Dog 99\%) over a weaker attack (Dog 60\%), mAP does not reflect the strength of an attack (see Fig. \ref{fig:mislabel}).
In addition, note that the overall detection error 
\begin{align}
\begin{split}
\mathcal{O}(x^{'}) - \mathcal{O}_{true} & = [\mathcal{O}(x^{'}) - \mathcal{O}(x)] + [\mathcal{O}(x) - \mathcal{O}_{true}] \\
& = \bm{\varepsilon}_{\textbf{\texttt{attack}}} + \varepsilon_{\texttt{model}},
\end{split}
\end{align}
where $\mathcal{O}_{true}$ is the ground truth model output, includes the attack error $\varepsilon_{\texttt{attack}}$ and the model error $\varepsilon_{\texttt{model}}$. The mAP measures the overall error by comparing the adversarial outputs with the ground truth, but we are only interested in the attack error $\varepsilon_{\texttt{attack}}=\mathcal{O}(x^{'}) - \mathcal{O}(x)$.
Therefore, we devise three new evaluation metrics. 
\begin{enumerate}
    \item Mean Confidence Variation: The average increase or decrease of the confidence value (before the sigmoid activation function) of all the bounding boxes at each iteration step $t$. This metric reflects the strength of the attack on the confidence value and is expressed as $\frac{1}{N}\sum_{i=1}^{N}{( c_{t,i} - c_{t-1,i} )}$.
    \item Number of Boxes: The total number of bounding boxes after the NMS. This metric shows how many objects are detected at each step of the attack $|\texttt{NMS}(O(x^{'}))|$.
    \item Relative Box Variation: After each iteration, the position of false positive bounding boxes fluctuates. This metric measures the percentage of consistent bounding boxes (bounding boxes that have the same position as in the previous step) at the current step and can be expressed as $\frac{|\texttt{NMS}(\mathcal{O}(x_t^{'}))| + |\texttt{NMS}(\mathcal{O}(x_{t-1}^{'}))| - |\texttt{NMS}(\mathcal{O}(x_{t}^{'}),\ \mathcal{O}(x_{t-1}^{'}))|}{|\texttt{NMS}(\mathcal{O}(x_t^{'}))|}$.
\end{enumerate}

Further, we compare our attack with the TOG attack \cite{chow2020adversarial}, which is also a gradient-based attack that updates the adversarial perturbation using the gradient of input with respect to the adversarial loss function. The main difference is that the TOG attack uses uniform initialization and updates the perturbation with a constant learning rate (no learning rate decay). Besides, the TOG attack uses the YOLO training loss \cite{redmon2018yolov3, bochkovskiy2020yolov4} as the adversarial loss function. We study the effect of these factors in the following sections, respectively.






\begin{figure*}[btph]
    \centering
    \includegraphics[width=0.80\linewidth]{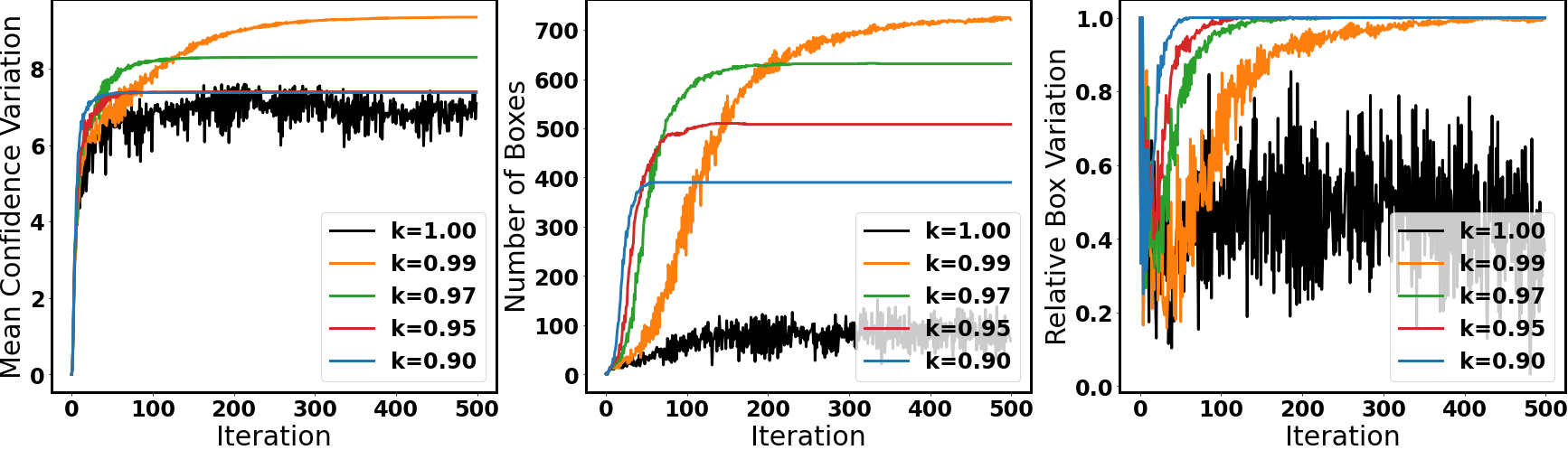}
    \caption{The  image-specific  PCB fabrication attack with different learning rate decays.}
    \label{fig:decay}
\end{figure*}

\begin{figure*}[btph]
    \centering
    \includegraphics[width=0.80\linewidth]{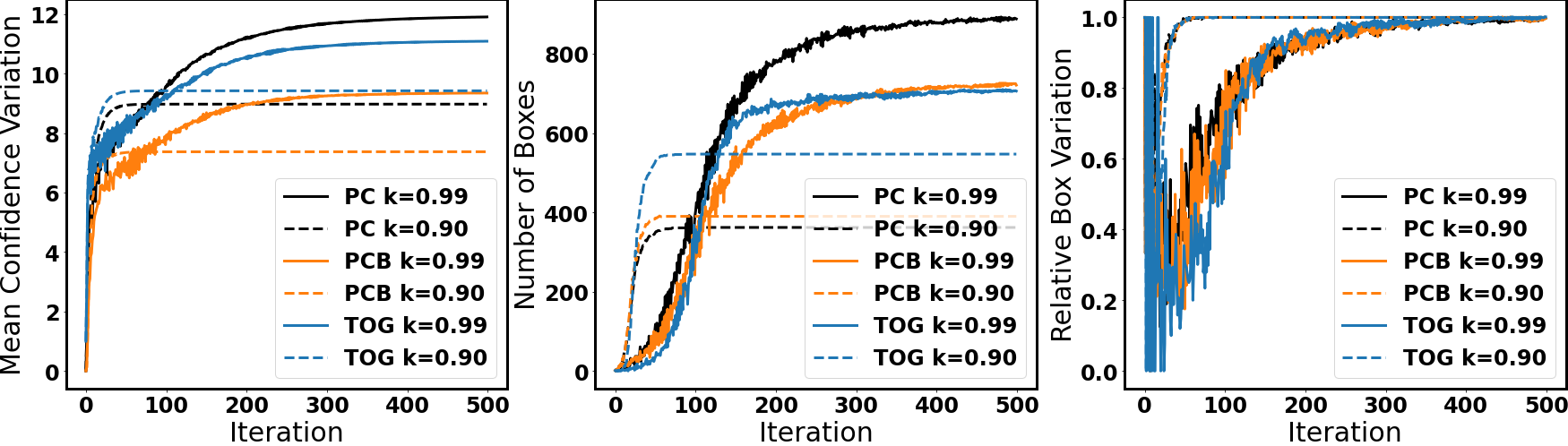}
    \caption{Different adversarial loss functions of the  image-specific  fabrication attack.}
    \label{fig:loss}
\end{figure*}

\begin{figure*}[tbp]
    \centering
    \begin{subfigure}[b]{0.22\textwidth}
        \centering
        \includegraphics[width=\linewidth]{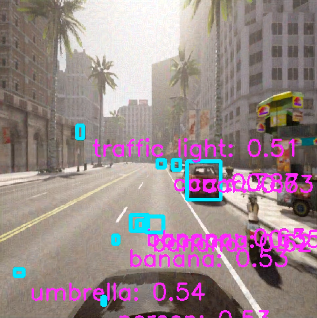}
        \caption{Clear Noon (City)}
        \label{fig:clear_noon} 
    \end{subfigure}
    \begin{subfigure}[b]{0.22\textwidth}
        \centering
        \includegraphics[width=\linewidth]{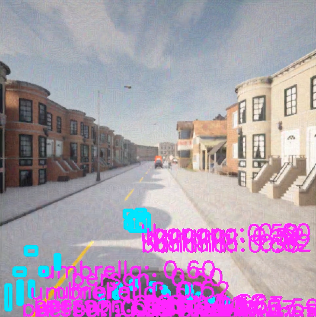}
        \caption{Clear Noon (Suburban)}
        \label{fig:clear_noon_suburban}
    \end{subfigure}
    \begin{subfigure}[b]{0.22\textwidth}
        \centering
        \includegraphics[width=\linewidth]{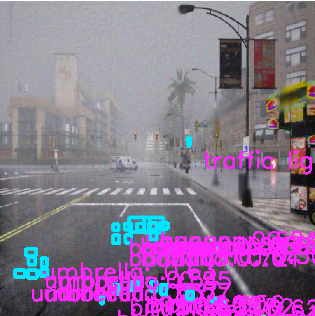}
        \caption{Hard Rain Sunset (City)}
        \label{fig:hard_rain} 
    \end{subfigure}
    \begin{subfigure}[b]{0.22\textwidth}
        \centering
        \includegraphics[width=\linewidth]{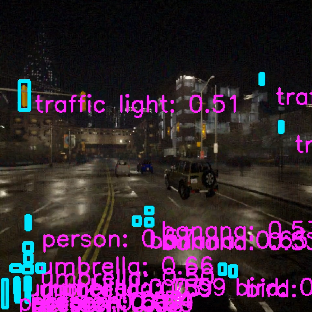}
        \caption{Wet Cloudy Night (City)}
        \label{fig:wet_cloudy}
    \end{subfigure}
  \caption{The UAP generalizes to different daytime, weathers and maps.}
  \label{fig:carla}
\end{figure*}

\subsection{Initialization Methods}

The TOG attack employs uniform initialization as noted in \cite{chow2020adversarial}. In contrast, other attacks use zero initialization, including \cite{fischer2017adversarial,li2021universal,madry2017towards}. Gradient-based attacks rely on gradients to iterate from the original image to an adversarial input. The uniform initialization may impede the initial gradient at the first iteration, potentially limiting the attack's effectiveness. To investigate this hypothesis, we conducted 10 runs of the attack with uniform initialization and compared the results with zero initialization using the mean confidence variation, number of boxes, and relative box variation.

Our results showed that, for the first two evaluation metrics, only two out of ten runs with uniform initialization resulted in a more effective attack than zero initialization. This supports our hypothesis that uniform initialization can impede the initial gradient and hinder the attack's effectiveness. Regarding the third evaluation metric, relative box variation, we observed convergence to 1 for both initialization methods, indicating that all false positive bounding boxes were stable, and the convergence speed was similar for both initialization methods.




\subsection{Learning Rate Decay}

To achieve a stable and efficient adversarial attack, it is necessary to avoid gradient counteraction, which can cause the attack to vary significantly over iterations. The PCB attack addresses this issue by introducing the learning rate decay factor $k$ to stabilize the attack over different iterations.

Neither the original PGD attack nor the TOG attack uses the learning rate decay and thus has an unstable iteration process (as shown by the black line in Fig. \ref{fig:decay}). This issue has not been thoroughly studied in prior research that relies on mAP as the evaluation metric. For example, at each step, we generate false positive bounding boxes at various positions, but none of them match the ground truth (mAP$=0$). As a result, even though the location of bounding boxes varies a lot (unstable) from one iteration to the next, the mAP stables at 0, which does not indicate the instability of the iteration process.

Using our new evaluation metrics, we can observe the complete iteration process (as depicted in Fig. \ref{fig:decay}). As the learning rate decay factor $k$ decreases from 0.99 to 0.90, more iterations are required before the two evaluation metrics, the mean confidence variation and the number of boxes, converge. However, both metrics converge at higher values when $k$ decreases, indicating a more efficient attack. 



\subsection{Adversarial Loss Function}

In this section, we aim to compare the effectiveness of three different adversarial loss functions for the fabrication attack (PC, PCB and TOG adversarial loss). Rather than determining the best loss function, our goal is to use our evaluation metrics to highlight the advantages and disadvantages of each method.


As shown in Fig. \ref{fig:loss}, we observe that the Relative Box Variation of all three methods converges to 1, indicating that the locations of most bounding boxes are stable in the final iterations. Though the PC attack generates the most bounding boxes and achieves the highest Mean Confidence Variation when $k=0.99$, it requires a larger number of iterations to reach the plateau, which can be computationally expensive when the number of sample images $X_{s}$ is large.

In contrast, both the TOG and PCB attacks converge faster and perform better than the PC attack when $k=0.90$. Therefore, we cannot definitively state that one method is superior to the others. Instead, the newly proposed evaluation metrics provide useful references for decision-making.


\begin{table*}[tpbh]
    \begin{subtable}[t]{\textwidth}
        \centering
        \begin{tabular}{lccca|ccca}
        \hline
        & \multicolumn{4}{c}{PCB Attack ($k=0.98$)} & \multicolumn{4}{c}{TOG Attack ($k=1.00$)}\\
        Iteration & 1 it & 10 it & 50 it & 100 it & 1 it & 10 it & 50 it & 100 it \\ 
        \hline
        \ \textbf{Map 01 (train)}  & \textbf{5.62} & \textbf{6.68} & \textbf{6.99} & \textbf{7.10} & \textbf{3.04} & \textbf{3.12} & \textbf{3.71} & \textbf{2.66} \\
        \ Map 02  & 2.09 & 3.23 & 3.65 & 3.82 & 0.84 & 0.76 & 1.04 & 0.75 \\
        \ Map 03  & 2.75 & 3.97 & 4.49 & 4.68 & 1.74 & 1.57 & 1.83 & 1.36 \\
        \ Map 04  & 4.30 & 5.37 & 5.85 & 6.05 & 2.71 & 2.40 & 2.75 & 2.23 \\
        \ Map 05  & 4.12 & 5.27 & 5.60 & 5.73 & 2.28 & 1.99 & 2.27 & 1.69 \\
        \ Map 06  & 4.48 & 5.63 & 6.04 & 6.18 & 2.76 & 2.55 & 2.95 & 2.47 \\
        \ Map 07  & 3.00 & 3.97 & 4.39 & 4.55 & 2.04 & 1.87 & 2.31 & 1.88 \\
        \hline
        \end{tabular}
        \caption{The mean confidence variation.}
        \label{tab:trans_conf}
        \vspace{0.05in}
    \end{subtable}
\hspace{\fill}
    \begin{subtable}[t]{\textwidth}
        \centering
        \begin{tabular}{lccca|ccca}
        \hline
        & \multicolumn{4}{c}{PCB Attack ($k=0.98$)} & \multicolumn{4}{c}{TOG Attack ($k=1.00$)}\\
        Iteration & 1 it & 10 it & 50 it & 100 it & 1 it & 10 it & 50 it & 100 it \\ 
        \hline
        \ \textbf{Map 01 (train)}  & \textbf{165.83} & \textbf{280.85} & \textbf{323.39} & \textbf{346.00} & \textbf{3.88} & \textbf{1.71} & \textbf{1.99} & \textbf{2.70} \\
        \ Map 02  & 1.89 & 21.72 & 49.71 & 63.54 & 1.61 & 1.47 & 1.61 & 1.63 \\
        \ Map 03  & 4.68 & 10.47 & 18.45 & 24.78 & 2.20 & 2.05 & 2.18 & 2.16 \\
        \ Map 04  & 32.29 & 74.44 & 87.96 & 100.46 & 0.78 & 0.71 & 1.32 & 0.74 \\
        \ Map 05  & 1.66 & 38.44 & 59.73 & 70.03 & 1.08 & 1.10 & 1.20 & 0.98 \\
        \ Map 06  & 12.80 & 41.05 & 52.57 & 66.79 & 3.41 & 3.02 & 3.47 & 3.17 \\
        \ Map 07  & 6.25 & 27.53 & 30.61 & 35.16 & 0.69 & 0.66 & 0.77 & 0.70 \\
        \hline
        \end{tabular}
        \caption{The average number of boxes.}
        \label{tab:trans_box}
        \vspace{0.05in}
    \end{subtable}
\hspace{\fill}
    \begin{subtable}[t]{\textwidth}
        \centering
        \begin{tabular}{lccca|ccca}
        \hline
        & \multicolumn{4}{c}{PCB Attack ($k=0.98$)} & \multicolumn{4}{c}{TOG Attack ($k=1.00$)}\\
        Iteration & 1 it & 10 it & 50 it & 100 it & 1 it & 10 it & 50 it & 100 it \\ 
        \hline
        \ \textbf{Map 01 (train)}  & \textbf{0.19\%} & \textbf{0.11\%} & \textbf{0.09\%} & \textbf{0.08\%} & \textbf{36.43\%} & \textbf{43.23\%} & \textbf{32.32\%} & \textbf{43.33\%} \\
        \ Map 02  & 60.65\% & 6.40\% & 2.27\% & 1.46\% & 80.37\% & 81.92\% & 86.92\% & 83.50\% \\
        \ Map 03  & 63.88\% & 24.64\% & 14.34\% & 11.16\% & 87.34\% & 86.02\% & 85.99\% & 87.41\% \\
        \ Map 04  & 2.60\% & 0.54\% & 0.44\% & 0.31\% & 68.17\% & 66.50\% & 54.42\% & 64.89\% \\
        \ Map 05  & 51.92\% & 1.30\% & 0.47\% & 0.30\% & 81.33\% & 76.89\% & 85.33\% & 76.00\% \\
        \ Map 06  & 24.90\% & 4.30\% & 2.37\% & 1.50\% & 95.29\% & 96.59\% & 96.79\% & 96.61\% \\
        \ Map 07  & 32.55\% & 1.87\% & 1.38\% & 0.98\% & 75.14\% & 73.78\% & 74.89\% & 76.72\% \\
        \hline
        \end{tabular}
        \caption{The relative box variation (the percentage of consistent bounding boxes).}
        \label{tab:trans_rela_box}
    \end{subtable}
\hspace{\fill}
    \caption{Evaluation results of the image-agnostic attack tested on the CARLA dataset ($\epsilon=8$).}
    \label{tab:trans}
\end{table*}

\subsection{The Attack Performance and Transferability}

We used an autonomous driving dataset collected from the CARLA simulator \cite{deschaud2021kitticarla}. The dataset includes driving records from 7 maps, including different areas (city, rural, urban, highway), daytime (noon, sunset, night), and weather (clear, cloudy, soft rain, and hard rain) (see Fig. \ref{fig:carla}). A 30-second driving video was collected from each map, sampling at 10 FPS, which is the same as the KITTI autonomous vehicle. 

For the image-specific PCB attack, we measured the performance of the attack on an NVIDIA RTX 2080Ti GPU. The image-specific attack achieved 12.44 FPS, faster than the 10 FPS sampling rate of the KITTI autonomous vehicle, thus achieving a real-time attack.

For online attacks against video streams, it is unnecessary to re-generate the adversarial perturbation for each frame because there is a high correlation between consecutive video frames. Thus, we can reuse the perturbation computed from the previous frame and iterate only one step for each frame to achieve real-time performance. 


For the image-agnostic attack, we trained the UAPs using Map 01. The same attack strength ($\epsilon=8$) is used for both PCB and TOG attacks. Since the original TOG attack does not use learning rate decay, we set $k=1.00$. After each iteration, we tested the attack performance on all seven maps using the three evaluation metrics (see Tab. \ref{tab:trans}).

\begin{enumerate}
    \item The mean confidence variation reflects the changes in the model outputs (before softmax) after applying the UAP. The experimental results show that our image-agnostic PCB attack produces higher variations than TOG, which indicates a stronger UAP. Besides, our method has a more stable iteration process, while the TOG generates the UAP in a more stochastic way (similar to Fig. \ref{fig:decay}).
    \item The average number of boxes measures the total number of bounding boxes after applying the UAP. The image-agnostic attack generates the most number of boxes on the training map but also generalizes to other maps. With learning rate decay, the PCB attack generates significantly more bounding boxes than the TOG attack.
    \item The relative box variation computes the percentage of consistent bounding boxes. During the UAP training process, as the UAP becomes stable, this metric converges to 100\%, meaning all bounding boxes are consistent. However, to measure the final attack performance, a high value (100\%) indicates the model generates the same bounding boxes before and after applying the UAP. Thus, a successful attack has a low value (0\%), indicating none of the bounding boxes are the same as the ones predicted without applying the UAP.
\end{enumerate}

\begin{figure}[tbhp]
    \centering
    \includegraphics[width=\linewidth]{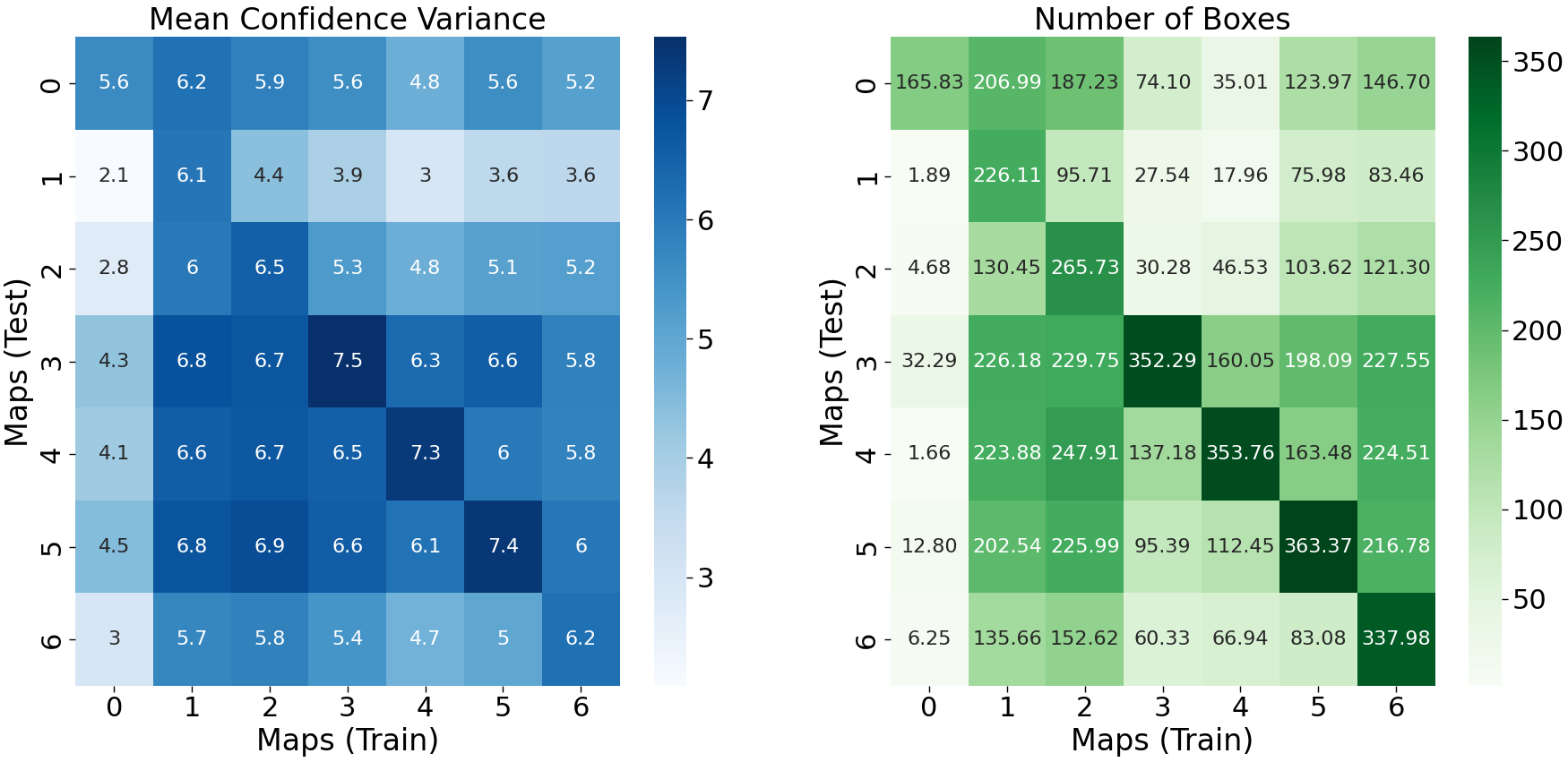}
  \caption{The heatmap of UAPs trained on seven maps.}
  \label{fig:cor}
\end{figure}

We also measured the transferability of the UAPs across maps (100 iterations, $k=0.98, \epsilon=8$). The image-agnostic PCB attack achieves the best performance on the training map but also generalizes to other maps (see Fig. \ref{fig:cor}). In summary, experimental results using our evaluation metrics demonstrate the importance of learning rate decay for UAP training.


\section{Discussion}

In real-world applications, the operating system that deploys the deep learning models is highly secure, while sensors are exposed to the environment to collect the data. Besides eavesdropping the sensor data, adversaries can inject the UAPs directly into the camera. In other words, the manufacturer can design adversarial cameras that bypass specific object detection models. Our research intends to call attention to the security of the sensor for deep learning applications. 


\clearpage

\section{Conclusions}


This paper presents a novel hardware attack that reveals a previously unknown vulnerability of deep learning object detection systems, posing serious threats to safety-critical applications. Unlike existing attack frameworks, our approach does not rely on any assumption about access to the object detection system, but rather leverages perturbations injected at the hardware level. Our experiments on the VOC2012 dataset, the CARLA dataset, and the YOLO detection model demonstrate the high efficiency of our attack. Further research may explore the extension of the attack to other tasks beyond object detection, or to other sensors, such as Lidar.

\bibliographystyle{IEEEtran}
\bibliography{IEEEabrv, mybibfile}

\end{document}